\documentclass[11pt,a4paper]{article}
\usepackage[hyperref]{emnlp-ijcnlp}
\usepackage{times}
\usepackage{latexsym}
\usepackage{amsmath}
\usepackage{amsfonts}
\usepackage{amsthm}
\usepackage{url}
\usepackage{tikz}
\usepackage{subcaption}
\usepackage{xspace}
\usepackage{adjustbox}
\usepackage{booktabs}
\usepackage{mathtools}
\usepackage{xfrac}
\usepackage{xspace}
\usepackage{comment}
\usepackage{microtype}
\usepackage{multirow}
\usepackage{algorithm}
\usepackage[noend]{algpseudocode}

\usepackage[disable]{todonotes}
\makeatletter
\newcommand*\iftodonotes{\if@todonotes@disabled\expandafter\@secondoftwo\else\expandafter\@firstoftwo\fi}  
\makeatother

\newcommand{\embed}{\mathbf{e}}
\newcommand{\defn}[1]{}

\aclfinalcopy 

\usepackage{color}

\usepackage{cleveref}
\crefname{section}{\S}{\S\S}
\Crefname{section}{\S}{\S\S}
\crefname{table}{Table}{}
\crefname{figure}{Figure}{}
\crefname{algorithm}{Algorithm}{}
\crefname{equation}{eq.}{}
\crefname{appendix}{App.}{}
\crefformat{section}{\S#2#1#3}  

\newcommand{\topic}[1]{}

\newcommand{\vg}{{\boldsymbol {g}}}

\newcommand{\msc}{\textsc{msc}}
\newcommand{\fem}{\textsc{fem}}

\newcommand{\calG}{{\cal G}}

\newcommand{\word}[1]{\textit{#1}}
\newcommand{\XX}{18\xspace} 
\newcommand{\YY}{5\xspace} 
\newcommand{\ZZ}{90\xspace} 
\newcommand{\genderList}{Bulgarian, Catalan,  Greek, Spanish, French, Hebrew, Hindi, Croatian, Italian, Lithuanian, Latvian, Polish, Portuguese, Romanian, Russian, Slovak,  Slovenian, and Ukranian\xspace} 

\newcommand{\genderlessList}[1][]{English, Japanese, Korean, Mandarin Chinese, %
\ifthenelse{\equal{#1}{}}{and }{or }%
Turkish}

\usepackage{linguex}

\usepackage[safe]{tipa}

\aclfinalcopy 

\title{Quantifying the Semantic Core of Gender Systems}
\author
  {
	\begin{tabular}{lllll}
	Adina Williams\raise1.0ex\hbox{\normalfont\normalsize \textschwa}\raise1.0ex\hbox{\normalfont \normalsize} & Ryan Cotterell\raise1.0ex\hbox{\normalfont\normalsize \textipa{S,H}}\raise1.0ex\hbox{\normalfont\normalsize} & Lawrence Wolf-Sonkin\raise1.0ex\hbox{\normalfont\normalsize \textipa{S}}
	\end{tabular} \\
		\begin{tabular}{lllll}
 \textbf{Dami{\'a}n Blasi}\raise1.0ex\hbox{\normalfont\normalsize \textipa{P}}
	& \textbf{Hanna Wallach}\raise1.0ex\hbox{\normalfont\normalsize \textipa{Z}}
	\end{tabular}
	\\
    \raise1.0ex\hbox{\normalfont\normalsize \textschwa} Facebook AI Research, New York, USA \\
    \raise1.0ex\hbox{\normalfont\normalsize \textipa{S}}Department of Computer Science, Johns Hopkins University, Baltimore, USA \\
    \raise1.0ex\hbox{\normalfont\normalsize \textipa{H}}Department of Computer Science and Technology, University of Cambridge, Cambridge, UK\\
     \raise1.0ex\hbox{\normalfont\normalsize \textipa{P}}Comparative Linguistics Department, University of Z\"{u}rich, Switzerland \\
    \raise1.0ex\hbox{\normalfont\normalsize \textipa{Z}}Microsoft Research, New York City, USA \\
	{\tt {adinawilliams@fb.com, rdc42@cam.ac.uk, lawrencews@jhu.edu}} \\
	{\tt {damian.blasi@uzh.ch, hanna@dirichlet.net}}
}

\begin{document}
\maketitle
\begin{abstract}
Many of the world's languages employ grammatical gender on the lexeme. For example, in Spanish, the word for house (\textit{casa}) is feminine, whereas the word for paper (\textit{papel}) is masculine. To a speaker of a genderless language, this assignment seems to exist with neither rhyme nor reason. But is the assignment of inanimate nouns to grammatical genders truly arbitrary? We present the first large-scale investigation of the arbitrariness of noun--gender assignments. To that end, we use canonical correlation analysis to correlate the grammatical gender of inanimate nouns with an externally grounded definition of their lexical semantics. We find that \XX languages exhibit a significant correlation between grammatical gender and lexical semantics.\looseness=-1
\end{abstract}

\section{Introduction}
In his semi-autobiographic work about his time traveling
through Germany, \textit{A Tramp Abroad}, \newcite{twaintramp} recounted his difficulty when learning the German gender system:
``Every noun has a gender, and there is no sense or system in
    the distribution; so the gender of each must be learned separately
    and by heart.
    In German, a young lady has no sex,
    while a turnip has. Think what overwrought reverence that shows
    for the turnip, and what callous disrespect for the girl.%
    ''
Although this humorous take on German grammatical gender is clearly a
caricature, the quote highlights the fact that the relationship between the grammatical gender of nouns and their lexical semantics is often quite opaque.
\looseness=-1

As arbitrary as certain noun--gender assignments may appear overall, a relatively clear relationship often exists between grammatical gender and lexical semantics for some of the lexicon. The portion of the lexicon where this relationship is clear usually consists of animate nouns; nouns referring to people morphologically reflect the sociocultural notion of ``natural genders.''\ This portion of the lexicon---the ``semantic core''---seems to be present in \emph{all} gendered languages \cite{aksenov1984,corbett1991gender}.
But how many \emph{inanimate} nouns can also be included in the semantic core? Answering this question requires investigating whether there is a correlation between grammatical gender and lexical semantics for inanimate nouns.\looseness=-1

Our primary technical contribution is demonstrating that grammatical
gender and lexical semantics can be correlated using canonical
correlation analysis (CCA)---a standard method for computing the
correlation between two multivariate random variables. We consider \XX
gendered languages, following 3 steps for each: First, we encode
each inanimate noun as a one-hot vector representing the noun's
grammatical gender in that language; we then create \YY
operationalizations of each noun's lexical semantics using word
embeddings in \YY genderless ``donor'' languages (\genderlessList[]); finally, for each genderless language, we use CCA to compute
the desired correlation between grammatical gender and lexical
semantics. This process yields a single value for each of the \ZZ
gendered--genderless language pairs, revealing a significant
correlation between grammatical gender and lexical
semantics for 55 of these language pairs.\looseness=-1

Secondarily, we investigate semantic similarities between the \XX
languages' gender systems---i.e.,\ their assignments of nouns to
grammatical genders. We analyze the projections of lexical semantics
(operationalized as word embeddings in English) obtained via CCA,
finding that phylogenetically similar languages have more similar
projections.\looseness=-1

\section{Background and Assumptions}

\subsection{Grammatical Gender}

Languages range from employing no grammatical gender on inanimate
nouns, like \genderlessList[], to drawing grammatical
distinctions between tens of gender-like
classes~\cite{corbett1991gender}. Although there are many theories
about the assignment of inanimate nouns to grammatical genders, to the
best of our knowledge, the linguistics literature lacks any
large-scale, quantitative investigation of arbitrariness of
noun--gender assignments. However, with the advent of modern
NLP methods---particularly with advancements in distributional approaches to
semantics~\citep{harris1954,firth1957}---and with the copious amounts of
text available on the internet, it is now possible to conduct such an
investigation. We focus on languages that have either two
(masculine--feminine) or three (masculine--feminine--neuter) genders,
to which nouns are exhaustively assigned, and investigate whether a correlation exists between
grammatical gender and lexical semantics for inanimate nouns---i.e.,\
whether noun--gender assignments are arbitrary or not.

In many languages, a noun's grammatical gender can be predicted from
its spelling and pronunciation~\citep{cucerzan2003, nastase2009}.
For example, almost all Spanish nouns ending in \word{-a} are
feminine, whereas Spanish nouns ending in \word{-o} are usually
masculine. These assignments are non-arbitrary; indeed,
\newcite[Ch.\ 4]{corbett1991gender} provides a thorough typological
description of how phonology pervades gender systems. We emphasize
that these assignments are not the subject of our
investigation. Rather, we are concerned with the relationship between
grammatical gender and lexical semantics---i.e.,\ when asking why the
Spanish word \word{casa} is feminine, we do not consider that it ends
in \word{-a}.\looseness=-1

Finally, our investigation is related to that of \citeauthor{kann},
which assumes that noun--gender assignments are non-arbitrary and
examines the predictability of grammatical gender from lemmatized word
embeddings; in contrast, we investigate the arbitrariness of
noun--gender assignments.

\begin{table*}[t!]
\begin{center}
\small
\begin{adjustbox}{max width=\linewidth}
\begin{tabular}{lcccccccccccccccccc}
\toprule
 & bg & ca  & el & es & fr & he & hi & hr & it & lt & lv & pl & pt & ro & ru & sk & sl & uk \\
\midrule
en & \bf 2596 & \bf 2720 & \bf 2872 & \bf 4947 & \bf 6257 & \bf 1489 & 828 & 443 & \bf 4800 & \bf 923 & \bf 881 & \bf 1646 & \bf 3918 &  397 & \bf 5779 & 1056 & 293 & 975\\
ja & \bf  2586 & \bf 2596 & \bf 2886 & \bf 3383 &  \bf 4654 &  \bf 2223 &  \bf 1421 & 486 &  \bf 3849 & 1241 & \bf 1215 & \bf 1884 & \bf 3615 & \bf 497 & \bf 4532 & 375 & 419 & 1380 \\
ko & \bf  1856 &  \bf 1843  & 1982 &  1840 &  \bf 2812 &  \bf 1774 & 1269 & 371 &  \bf 2513 &  1089 &  \bf 997 & 1357 &  \bf 2442 &  \bf 364 &  \bf 2680 & 247 &  317 &   1169\\
tr &  \bf   1578 &  \bf  1623  &  \bf 1735 &  \bf  1766 &  \bf 2580 &  \bf 1275 & 817 & 274 &  \bf 2287 & 903 &  826 & 1163 &  \bf 2223 & 303 &  \bf 2470 & 218 & 281 & 909\\
zh &   \bf  2275 &  \bf 2190  &   2454 &  \bf 2693 &  \bf 3722 &  1810 &  \bf 1266 &  373 &  \bf 3170 &  \bf 1111 &  \bf 1110 &  \bf 1643 &  \bf 3084 & 480 &  \bf 3652 & 286 &  \bf 341 & 1196\\
\bottomrule
\end{tabular}
\end{adjustbox}
\end{center}
\caption{The number of inanimate nouns for each gendered--genderless language pair. Bold indicates that our investigation reveals a significant correlation between grammatical gender and lexical semantics for that pair.}\label{tab:count}
\end{table*}
\subsection{Lexical Semantics via Word Embeddings}\label{sec:word-embeddings}

The NLP community has widely adopted word embeddings as way of
representing lexical semantics. The underlying motivation behind this
adoption is the observation that words with similar meanings will be
embedded as vectors that are closer together. As we explain in
\cref{sec:cca}, our investigation requires a definition of lexical
semantics that is independent of grammatical gender.\ However, in many
gendered languages, word embeddings effectively encode grammatical gender because
this information is trivially recoverable from distributional
semantics. For example, in Spanish, singular masculine nouns tend to
occur after the article \word{el}, whereas singular feminine nouns
tend to occur after the article \word{la}. For this reason, we use an
externally grounded definition of lexical semantics: we
create \YY operationalizations of each noun's lexical semantics using
word embeddings in \YY genderless ``donor'' languages (\genderlessList[]). We use \YY languages that are phylogenetically distinct and spoken in distinct regions to minimize any spurious correlations.\footnote{It is natural to ask whether polysemy and
  homonymy might result in spurious similarities. For example, in
  English, the words \word{fish$_N$} and \word{fish$_V$} are
  homonymous, but in Mandarin Chinese, the words \emph{y\"{u}}
  (\word{fish$_N$}) and \emph{di\`ao} (\word{fish$_V$}) are not. If
  patterns of homonymy in the genderless languages are very
  different, but patterns of correlation between grammatical gender
  and lexical semantics are very similar, then we can be reasonably
  sure that the correlations are not due to homonymy.}\looseness=-1

Our investigation is based on the linguistic assumption that word
embeddings in a genderless ``donor'' language are a good proxy
for genderless lexical semantics.\ In practice, however, this
assumption is generally false: word embeddings are largely a
reflection of the text with which they were trained.\ For example, the
embedding of the word \word{snow} will differ depending on whether the
training text was written by people near the equator or people near
the North Pole, even if both groups speak the same language. Such
differences will be more pronounced for rare words, which are arguably
more language- and culture-specific than many common words. For this
reason, we limit the scope of our investigation to only those inanimate nouns that are likely to be used consistently across different languages. To implement this limitation, we use a Swadesh
list~\citep{buck1949,swadesh1950,swadesh1952,swadesh1955, swadesh1971}---a list of words constructed to contain only very frequent words that are as close to culturally
neutral as possible.  By limiting the scope of our investigation to only those inanimate nouns that appear in a Swadesh list, we can be
reasonably confident that their word embeddings in \genderlessList[] are a good proxy for their genderless lexical semantics, as desired.\looseness=-1

\section{Methodology}\label{sec:cca}

\subsection{Data}

We use Open Multilingual WordNet \citep{elwordnet,hewordnet,hrwordnet2, frwordnet, rowordnet, plwordnet1, bgwordnet, itwordnet, bond2012,  slwordnet, cawordnet, powordnet,  plwordnet2, bond2013, ltwordnet,  hrwordnet1}\footnote{\href{http://compling.hss.ntu.edu.sg/omw/summx.html}{http://compling.hss.ntu.edu.sg/omw/summx.html}} as our Swadesh list. This yields \XX gendered languages (\genderList), of which 17 are from the Indo-European family and 1 are not.\footnote{This imbalance as a limitation of our investigation.\looseness=-1} For the 5 genderless languages (\genderlessList[]), we use pre-trained, 50-dimensional word embeddings from \textsc{fastText}
\cite{bojanowski2017enriching, grave2018}.\footnote{The
  \textsc{fastText} word embeddings were trained using Common Crawl
  and Wikipedia data, using CBOW with position weights, with character
  $n$-grams of length 5. For more information, see
  {\href{fasttext.cc/docs/en/crawl-vectors.html}{http://fasttext.cc/docs/en/crawl-vectors.html}}.} For each gendered--genderless language pair, we limit the scope of our investigation to only those inanimate nouns that occur in both our Swadesh list and in \textsc{fastText}; we provide the resulting counts in Table \ref{tab:count}. Finally, we randomly partition the set of nouns for each language pair into a 75\%--25\% training--testing split.\looseness=-1

\subsection{Notation}

We first establish the requisite notation. Let
$\mathcal{V}_{\ell, m} = \{1,\ldots, V_{\ell,m}\}$ denote a set of
integers representing the inanimate nouns for gendered language $\ell$
and genderless language $m$. Let $\calG_\ell$ denote the
(arbitrarily ordered) genders in language $\ell$; for example, let $\calG_\textit{spanish} =
(\msc, \fem)$. Given an inanimate noun $n \in \mathcal{V}_{\ell,m}$, let $\vg_\ell(n)$
denote a one-hot vector representing $n$'s grammatical gender in
language $\ell$, so that the $i^\textrm{th}$ entry corresponds to the $i^\textrm{th}$ gender in $\calG_\ell$. Similarly, let $\embed_m(n) \in \mathbb{R}^{50}$ denote the
50-dimensional word embedding
representing the lexical semantics of $n$ in language $m$. Let $G_\ell \in
\mathbb{R}^{|\calG_{\ell}| \times V_{\ell,m}}$ collectively denote the
inanimate nouns' grammatical genders in language $\ell$, so that the
$n^\textrm{th}$ column is $\vg_\ell(n)$, and let $E_m \in \mathbb{R}^{50
  \times V_{\ell,m}}$ collectively denote the inanimate nouns' lexical
semantics, so that the $n^\textrm{th}$ column is $\embed_m(n)$. Finally,
let $G^{\textrm{train}}_\ell$ and $E^{\textrm{train}}_m$ respectively denote the columns of $G_\ell$ and $E_m$ that correspond to the inanimate nouns in the training set and let $G^{\textrm{test}}_\ell$ and $E^{\textrm{test}}$ respectively denote the columns of $G_\ell$ and $E_m$ that correspond to the
inanimate nouns in the testing set.\looseness=-1

\subsection{Canonical Correlation Analysis}

CCA is a standard method for computing the correlation between two
multivariate random variables. In our investigation, we are interested
in the correlation between grammatical gender and lexical semantics
for each gendered--genderless language pair. To compute this
correlation, we start by solving the following optimization
problem:\looseness=-1
\begin{equation*}\label{eq:cca}
(\mathbf{a}^\star, \mathbf{b}^\star) = \underset{(\mathbf{a}, \mathbf{b})}{\text{arg\,max}} \,\,\textit{corr}(\mathbf{a}^{\top} G^{\textrm{train}}_\ell, \mathbf{b}^{\top} E_m^{\textrm{train}}).
\end{equation*}
\begin{figure}[b!]
   \centering
    \includegraphics[angle=-90, width=\columnwidth]{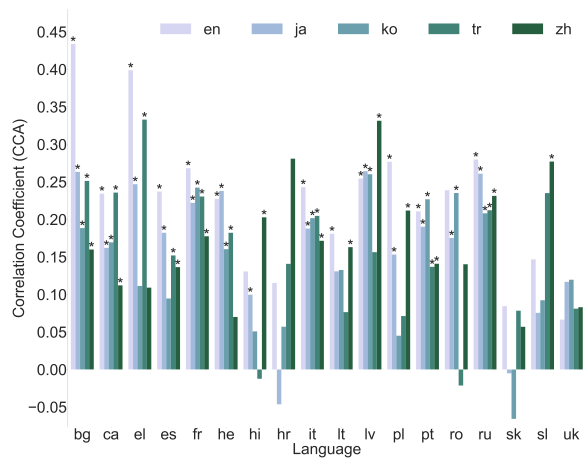} 
    \caption{The correlation between grammatical gender and lexical semantics for each of the 90 gendered--genderless language pairs ($*$ indicates significance).\looseness=-1}
    \label{fig:spearman}
\end{figure}
Although this optimization problem is non-convex, it can be solved in
closed form using singular value decomposition (SVD). We use a
standard implementation of CCA~\cite{pedregosa2011scikit}.\looseness=-1

Having found the projections $\mathbf{a}^\star \in
\mathbb{R}^{|\mathcal{G}_\ell|}$ and $\mathbf{b}^\star \in
\mathbb{R}^{50}$ that maximize the correlation, we then use them to
compute the correlation between grammatical gender and lexical
semantics as follows:
\begin{equation*}
  \rho_{\ell,m} = \textit{corr}({\mathbf{a}^{\star}}^\top
G^{\textrm{test}}_\ell, {\mathbf{b}^\star}^{\top}
E_m^{\textrm{test}}).
\end{equation*}%

To establish statistical significance, we follow the approach
of~\citet{monteiro16multiple}. We create $B\!=\!100,000$ permutations of the
columns of $G^{\textrm{train}}_\ell$; for each permutation $b$, we
then repeat the steps above to obtain $\rho_{\ell,m}$;
finally, we compute\looseness=-1
\begin{equation*}
  p = \frac{1 + \sum_{b=1}^{B} \delta (\rho^b_{\ell,m} \geq \rho_{\ell,m})}{B+1}.
\end{equation*}
Because our investigation involves testing 90 different hypotheses, we
use Bonferroni correction~\cite{dror-etal-2017-replicability}---i.e.,\
we multiply $p$ by 90. If the resulting Bonferroni-corrected $p$-value
is small, then we can reject the null hypothesis that there is no
correlation between grammatical gender and lexical semantics for that
language pair.\looseness=-1

Secondarily, we investigate semantic similarities between the 18
languages' gender systems by analyzing their projections of lexical
semantics. For each pair of \emph{gendered} languages $\ell$ and
$\ell'$, we compute the correlation (cosine distance) between
$\mathbf{b}^{\star}_\ell$ and $\mathbf{b}^{\star}_{\ell'}$ for each of
the 5 genderless languages. \looseness=-1



\section{Results}

We find a significant correlation between grammatical gender and
lexical semantics (i.e.,\ the Bonferroni-corrected $p$-value is less
than 0.05) for 55 of the \ZZ gendered--genderless language
pairs. These results are depicted in \cref{fig:spearman}. For Slovak, Croatian, and Ukranian, we find no correlation for any of the genderless languages; for
Slovenian, we find a significant correlation for only
Mandarin Chinese. We suspect that these results
are due the relatively small number of inanimate nouns considered for
each of these language pairs (see Table \ref{tab:count} for the counts). We also
find slightly different patterns of correlation for the different
genderless languages that we use to create our 5 operationalizations
of lexical semantics. For Japanese, we
find significant correlations for 13 of the 18 gendered languages; for English and Chinese, we find significant correlations for 12; for
Korean and Turkish, we find significant correlations for 9 of the gendered
languages.\looseness=-1

\begin{figure}[t!]
\vspace{-1cm}
    \includegraphics[width=\columnwidth]{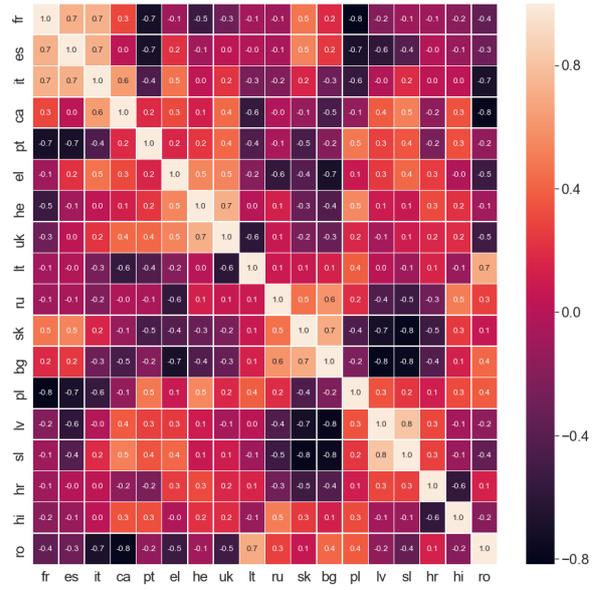}\\
\vspace{-2.1cm}\\
    \caption{The correlation between
      $\mathbf{b}^{\star}_\ell$ and $\mathbf{b}^{\star}_{\ell'}$ for each pair of gendered languages $\ell$ and $\ell'$ (for English).} \vspace{-0.41cm}
    \label{fig:dendorgram}
\end{figure}

For each pair of gendered languages $\ell$ and $\ell'$, Figure
\ref{fig:dendorgram} depicts the the correlation (cosine distance)
between $\mathbf{b}^{\star}_\ell$ and $\mathbf{b}^{\star}_{\ell'}$ for
English. We find higher correlations for pairs of languages that are phylogenetically similar. For example, French has higher correlations with Spanish and Italian than with Polish. This is likely because phylogenetically
similar languages exhibit historical similarities in their gender systems as a result of a common linguistic origin
\citep{fodor1959,ibrahim2014, stump2015}.\looseness=-1

\section{Conclusion}

Our investigation is the first to quantitatively demonstrate that
there is a significant correlation between grammatical gender and
lexical semantics for inanimate nouns. Although our results provide
evidence for the non-arbitrariness of noun--gender assignments, they
must be contextualized. In contrast to animate nouns, it is not clear
that a single cross-linguistic category explains our results. Moreover, we
limit the scope of our investigation to frequent inanimate
nouns. These nouns tend to be distributed across genders, whereas less
frequent inanimate nouns tend to be assigned to a single
gender~\citep{Dye2015AFT}. We leave the investigation of less frequent
inanimate nouns for future work.\looseness=-1

\section*{Acknowledgements}
Thanks to Jennifer Culbertson, Jacob Eisenstein, and Arya McCarthy for conversations on this topic.

\bibliography{naaclhlt2019}
\bibliographystyle{acl_natbib}

\end{document}